\makeatletter
\def\input@path{{DSArena/}}
\makeatother
\PassOptionsToPackage{table}{xcolor}
\ifdefined\pdfminorversion\pdfminorversion=7\fi
\documentclass[10pt]{fairmeta}
\usepackage[T1]{fontenc}
\usepackage{amsmath,amssymb,amsthm,booktabs,array,tabularx,longtable}
\usepackage{newtxtext,newtxmath}
\usepackage[table]{xcolor}
\usepackage[numbers,square,sort&compress]{natbib}

\usepackage{microtype,enumitem,float,placeins}
\usepackage[colorlinks=true,linkcolor=metablue,citecolor=metablue,urlcolor=metablue]{hyperref}
\definecolor{metablue}{HTML}{A51C30}
\definecolor{metabg}{HTML}{F8EEF0}
\definecolor{metafg}{HTML}{353B42}
\definecolor{arenateal}{HTML}{568A91}
\definecolor{tablehead}{HTML}{F3F1EB}
\hypersetup{pdftitle={DSArena: Evaluating LLMs as Data Science Agents through Human Preferences},pdfsubject={Real human preference evaluation with Bradley-Terry ranking},pdfkeywords={DSArena, data science agents, human preference, Bradley-Terry, ranking}}
\renewcommand{\fairinstitutionlogos}{%
  \includegraphics[height=0.72cm]{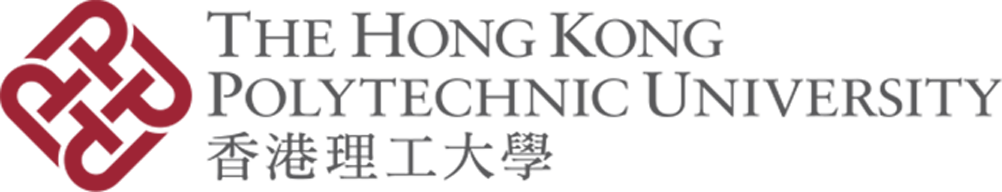}%
  \hspace{10pt}%
  \includegraphics[height=0.8cm,trim=64bp 30bp 24bp 18bp,clip]{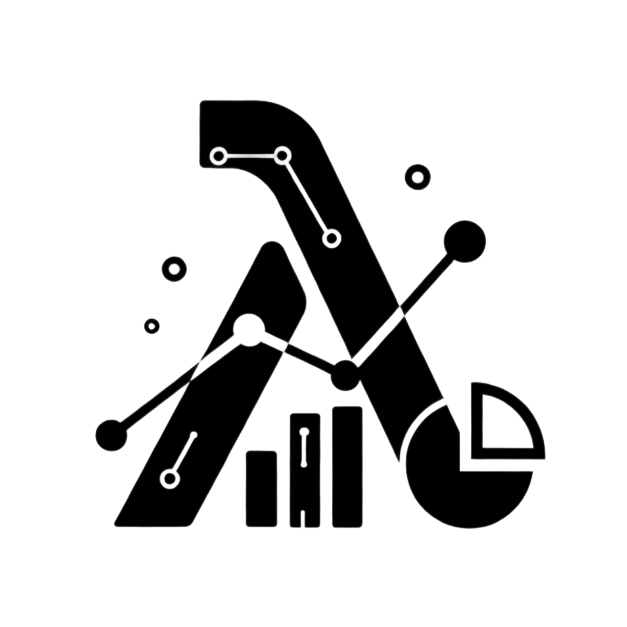}%
}
\fancypagestyle{fairfirstpage}{%
  \fancyhf{}%
  \fancyhead[L]{\raisebox{-\dp\strutbox}{\fairinstitutionlogos}}%
}
\renewcommand{\arraystretch}{1.14}
\newcommand{\R}{\mathbb R}

\newcommand{\argmin}{\operatorname*{arg\,min}}

\title{DSArena: Evaluating LLMs as Data Science Agents through Human Preferences}
\author[1]{Maojun Sun}
\author[1,\dagger]{Jian Huang}
\author[1,\dagger]{Ruijian Han}
\author[1,\dagger]{Yancheng Yuan}
\affiliation[1]{The Hong Kong Polytechnic University}
\contribution[\dagger]{Joint corresponding author}
\metadata[LAMBDA]{\href{https://lambda.com.ai/}{lambda.com.ai}}
\metadata[DSArena Leaderboard]{\href{https://lambda.com.ai/dsarena}{lambda.com.ai/dsarena}}
\abstract{Large language model (LLM)-based data science agents show strong potential for automating end-to-end analytical workflows, yet evaluating their performance remains challenging because real-world problems are open-ended and rarely have a single correct solution. Existing benchmarks rely largely on rule-based checks or LLM judges, capturing only a partial view of what users value. We introduce DSArena, an online platform for evaluating LLMs as data science agents through human preferences. Users submit their own tasks and datasets and compare two complete analyses in randomized A/B positions, with model identities hidden until voting. Applying Bradley--Terry estimation to 2,038 votes from 779 users across 12 LLMs, we find that Claude Opus 5 leads the leaderboard, followed by GPT-5.6 Sol and DeepSeek V4.1 Flash. Analyzing both subjective voting reasons and objective execution and report features shows that users most often cite completeness, while longer runtimes and more figures and tables consistently increase the odds of preference. Comparing human votes with three LLM judges on 500 sampled pairs reveals clear differences: humans select reports with more figures and pages 13.0\% and 11.3\% more often than the judge mean, respectively, whereas LLM judges more often favor more tool calls and generated files and emphasize validation, numerical consistency, and statistical support. These findings clarify human--AI evaluation gaps and provide preference data for reward modeling and agent training.}

\begin{document}
\maketitle

\begin{figure}[H]
\centering
\includegraphics[width=\linewidth]{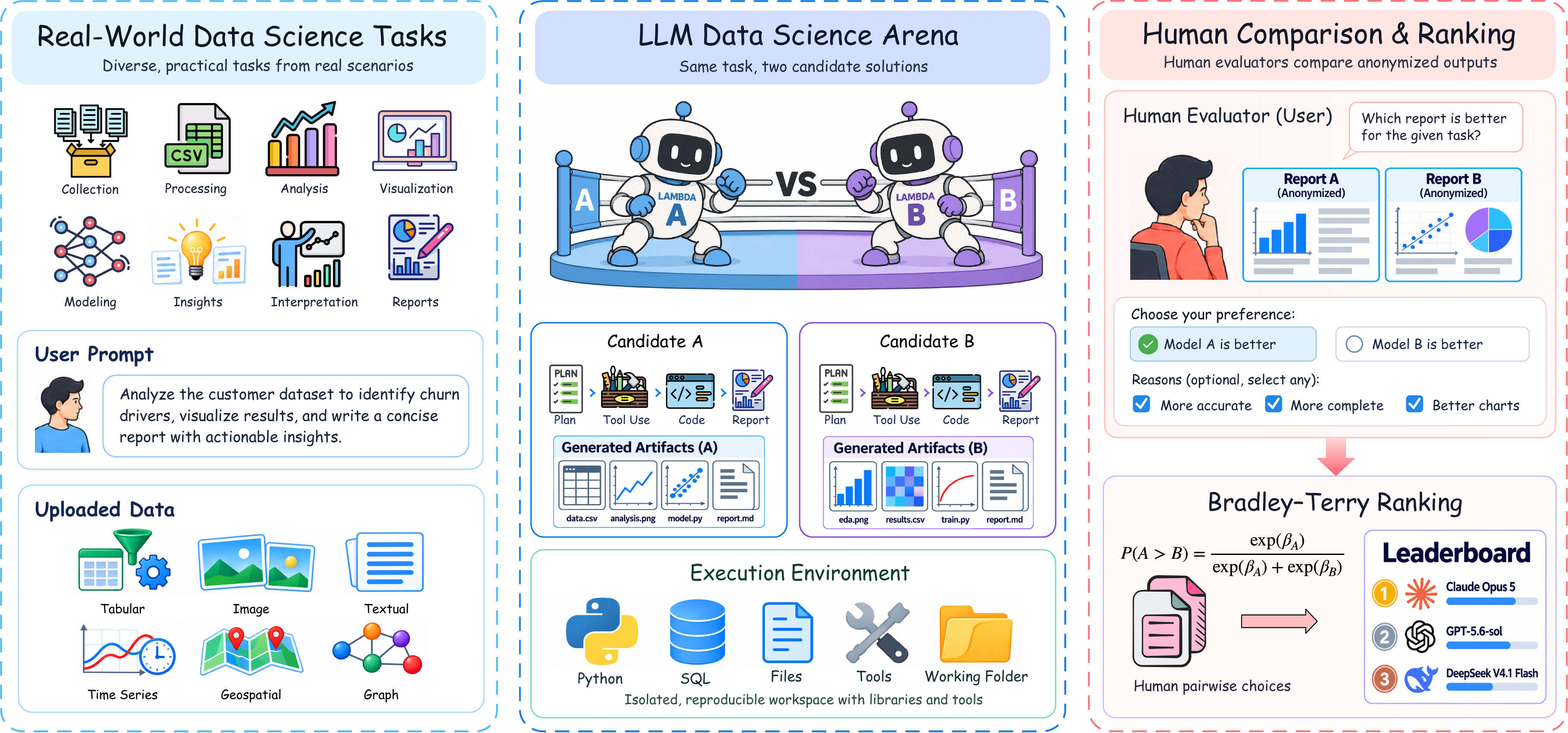}
\caption{Overview of DSArena. Two models analyze the same user task and dataset, and users compare their complete outputs in randomized A/B positions with model identities hidden until voting. Human votes drive the Bradley--Terry leaderboard, while subjective voting reasons and objective execution and report features support preference analysis.}
\label{fig:method}
\end{figure}

\section{Introduction}
Large language model (LLM)-based data science agents such as Data Interpreter, DeepAnalyze, and LAMBDA have shown strong potential for automating end-to-end data science workflows \citep{hong2025interpreter,zhang2025deepanalyze,sun2026lambda}. Given a natural-language request, these agents can prepare data, write and run code, fit statistical or machine-learning models, and generate analytical reports \citep{sun2026survey}. As their capabilities grow, a key challenge is how to evaluate the complete analyses they produce. Unlike closed-form coding tasks, real-world data science problems usually allow multiple valid approaches, and end-to-end agents respond with full reports combining numerical results, figures, tables, and conclusions. As a result, two analyses may both reach sound conclusions yet differ widely in clarity, chart quality, reproducibility, and practical usefulness to the user.

Existing benchmarks largely evaluate agents against fixed outputs and predefined targets, such as unit tests and API constraints for code generation \citep{lai2022ds1000}, closed-form answer checks and programmatic output verification \citep{hu2024infiagent,huang2024dacode,zhang2025datascibench}, or competition metrics for predictive modeling \citep{jing2024dsbench,chan2024mlebench}. While these protocols reliably test whether specific targets are met, they do not directly measure the coherence, completeness, or overall usefulness of a full analytical report from the user's perspective. Thorough evaluation therefore requires moving beyond checking whether code runs or whether a final answer matches a reference \citep{Donoho02012026,sun2026rejoinder1,sun2026rejoinder2}. In addition, static benchmarks carry a risk of data leakage, as their public datasets or reference solutions may appear in LLM training data.

To evaluate full analytical workflows, DSAEval uses LLM judges to assess reasoning, code, and final reports across multiple dimensions \citep{sun2026dsaeval}. Yet automated judgments do not necessarily match the diverse preferences of real users. In practice, preferences vary across people and tasks: depending on their goals, users weigh accuracy, clarity, completeness, and waiting time differently. To capture these real-world preferences and benchmark LLMs in live data science settings, we introduce \textbf{DSArena}, an online evaluation platform built on LAMBDA. On DSArena, users submit their own problems and datasets, inspect complete analyses generated by two competing LLMs, and select their preferred result through blinded pairwise voting.

Specifically, for each comparison, the user selects one model and the platform pairs it with a second model to solve the same task on the same data. The two resulting analyses are randomly assigned to positions A and B, and model names stay hidden until the vote is submitted, reducing both model-name and position bias. Alongside their choice, users can optionally record subjective voting reasons explaining their preference, while the platform logs objective execution statistics and report features for quantitative analysis. Figure~\ref{fig:method} illustrates the overall DSArena workflow. Because new user tasks, datasets, and models are continuously added, DSArena provides a live evaluation environment that reflects evolving model capabilities and real-world analysis needs.

Over four months, we collected 2,038 pairwise preference votes from 779 users across 12 LLMs, totaling 4,076 model appearances. We fit a Bradley--Terry (BT) model \citep{bradley1952rank} to estimate relative model strengths and build the leaderboard, and we examine what drives user preference from both subjective (voting reasons) and objective (execution statistics and report features) perspectives.

Our empirical results show that Claude Opus 5 leads the overall ranking, followed by GPT-5.6 Sol and DeepSeek V4.1 Flash. From the subjective perspective, users most often cite ``More complete'' as the reason for their choice, although different models lead across specific reason categories. From the objective perspective, longer execution times as well as higher figure and table counts show clear positive associations with user preference even after controlling for model identity and correlated features. Furthermore, comparing human votes with three LLM judges on 500 sampled comparisons reveals clear differences in evaluation priorities. The largest selection-rate gaps appear for figure count and report pages, where humans choose analyses with more figures and more pages $13.0\%$ and $11.3\%$ more often than the three-judge mean, respectively. In contrast, LLM judges more often select analyses with more tool calls and generated files, and their explanations emphasize validation, numerical consistency, and statistical support for conclusions. Together, these findings clarify both the report attributes associated with user choices and the different priorities of human and automated evaluators.

Our main contributions are summarized as follows:
\begin{enumerate}
\item \textbf{An online arena for real-world human evaluation.}
We build DSArena, an online platform for evaluating LLMs as data science agents through blinded pairwise human preferences over complete analyses of user-submitted tasks and datasets.

\item \textbf{A dynamic human-preference leaderboard.}
Over four months, we collect 2,038 votes from 779 users covering 12 LLMs, and we release a Bradley--Terry leaderboard that summarizes these preferences and updates as new votes and models arrive.

\item \textbf{Empirical analysis of human preferences in data science.}
We analyze what users value in complete analyses from both subjective and objective perspectives, combining user-reported voting reasons with objective execution statistics and report features. Our findings highlight the importance of analytical coverage, visual presentation, and the trade-off between depth and waiting time.

\item \textbf{Measuring human--LLM judge differences.}
On 500 sampled comparisons evaluated by three LLM judges, we show that humans more often favor visually rich reports, whereas LLM judges prioritize heavier tool use, validation, and statistical support for conclusions.
\end{enumerate}

\section{Related Work}
\subsection{LLMs as data science agents}
LLM-based data science agents automate workflows across data preparation, statistical analysis, modeling, and report generation \citep{sun2026dare}. Several system designs have been proposed for these multi-step pipelines. Data Interpreter represents task dependencies with hierarchical graphs and iteratively refines and verifies generated code \citep{hong2025interpreter}, whereas DS-Agent uses case-based reasoning to retrieve and adapt past solutions \citep{guo2024dsagent}. In competition workflows, AutoKaggle coordinates multiple specialized agents for data cleaning, feature engineering, and modeling, combining code execution with debugging and unit testing \citep{li2025autokaggle}. Taking an end-to-end training approach, DeepAnalyze trains an agentic LLM to carry out autonomous analysis from raw data to research reports \citep{zhang2025deepanalyze}. LAMBDA originally relied on cooperating programmer and inspector agents and has since evolved into a skill-guided, end-to-end data science agent framework \citep{sun2026lambda}. Together, these systems show the growing scope of automated data science and the need to evaluate the complete analyses delivered to end users. In this work, DSArena evaluates the LLMs powering these workflows within the unified LAMBDA framework.

\subsection{Evaluating data science agents}
Existing data science benchmarks mainly focus on code correctness, closed-form answers, or predictive performance. DS-1000 evaluates Python code generation across common data science libraries using execution tests and API constraints \citep{lai2022ds1000}. Moving toward multi-step analysis, InfiAgent-DABench turns analysis questions into closed-form outputs for automated grading \citep{hu2024infiagent}, DA-Code tests Python and SQL solutions in executable environments \citep{huang2024dacode}, and DSEval covers tasks across the data science lifecycle \citep{zhang2024dseval}. Similarly, DSBench includes realistic data analysis and modeling problems \citep{jing2024dsbench}, DataSciBench uses task-specific functions and programmatic rules to check execution outcomes \citep{zhang2025datascibench}, and DABstep evaluates multi-step industry analyses through automated checks on factoid-style answers \citep{egg2025dabstep}.

However, open-ended data science tasks typically produce full reports that combine methodology, numerical findings, visualizations, and conclusions. Evaluating these reports requires judging not only whether isolated outputs are correct, but also whether the overall analysis is clear, complete, and useful. DSAEval takes a step in this direction by using LLM judges to score complete analyses across reasoning, code, and final reports \citep{sun2026dsaeval}. Even so, its benchmark suite remains fixed, leaving it open to data contamination over time, and automated scoring offers only an indirect proxy for what real users value on their own tasks. DSArena complements static and judge-based benchmarks by evaluating models on live, user-submitted tasks through direct human preferences.

\subsection{LLM arenas for human preference evaluation}
Human-preference arenas evaluate models directly through side-by-side user choices in live settings. Chatbot Arena pioneered this approach by collecting blinded pairwise votes on conversational responses to build community-driven leaderboards \citep{chiang2024chatbot}. Later platforms extended the idea to specialized domains: Copilot Arena embeds pairwise comparisons into developers' coding environments to evaluate code completions \citep{chi2025copilot}, while SciArena applies community voting to long-form, literature-grounded scientific queries \citep{zhao2025sciarena}. Together, these platforms show how arena-based evaluation captures real user preferences in practice.

DSArena is the first platform to bring this arena approach to end-to-end data science. Unlike conversational replies, code completions, or literature summaries, data science tasks involve connected stages of data processing, feature engineering, modeling, and interpretation, producing a complete analytical report (for example, in PDF or PPTX format) rather than a single text response. This setting makes it possible to study human preferences over both the analysis process and the final report. By pairing blinded votes with subjective voting reasons and objective execution and report features, DSArena examines what users value in data science outputs and how those preferences vary across real-world tasks.

\subsection{Ranking models from pairwise preferences}

Pairwise ranking methods combine head-to-head outcomes against different opponents into a single scale of model strength. The Bradley--Terry model assigns each model a positive strength parameter and expresses its win probability against an opponent in terms of their relative strengths \citep{bradley1952rank}. Maximum-likelihood estimation combines observed comparisons into overall strength estimates and is commonly solved with iterative fitting algorithms, including minorization--maximization and Zermelo-type iterations \citep{hunter2004mm,han2026convergence}. To incorporate auxiliary information beyond baseline strengths, covariate-assisted Bradley--Terry models augment each item's intrinsic utility with feature effects \citep{schauberger2019btllasso}. Common formulations include static item-level covariates \citep{fan2024uncertainty} and comparison-specific dynamic covariates that capture contextual effects \citep{dong2026statistical}. DSArena uses regularized Bradley--Terry estimation to turn users' pairwise choices between data science analyses into a model leaderboard and adopts a dynamic-covariate formulation to control for position preference and evaluate how execution and report features associate with user preference. For a broader overview of recent advances in the Bradley--Terry model, see \citet{fang2026recent}.

\section{Methods}
\label{sec:methods}
Building on blinded pairwise votes, our analysis proceeds in four stages: aggregating votes into an overall leaderboard, analyzing subjective voting reasons, measuring associations with objective execution and report features, and comparing human choices with LLM judgments. We first define the tasks, agent workflows, and voting protocol shared across these analyses.

\subsection{Tasks, agents, and human preferences}
\label{sec:formulation}
Each pairwise comparison evaluates two complete analyses generated for the same user request, allowing users to judge both outputs directly against their own goals and data. Formally, we consider $n$ comparisons across $M$ LLM-based agents operating within LAMBDA. In comparison $i$, a user provides a natural-language instruction $p_i$ together with supporting data or files $\mathcal D_i$, defining the task $z_i=(p_i,\mathcal D_i)$. The user selects one model, and the platform pairs it with a second model to solve the same task. Appendix~\ref{app:notation} summarizes all notation in Table~\ref{tab:notation}.

Every agent runs inside LAMBDA's shared runtime, which provides memory, tools, skills, a unified action space, and an isolated sandbox \citep{sun2026lambda}. For agent $m$, its LLM parameters $\theta_m$ define a conditional action policy $\pi_{\theta_m}$. At step $t$, the agent samples an action $v_{imt}\sim\pi_{\theta_m}(\cdot\mid z_i,H_{imt})$, where the context $H_{imt}$ contains the dialogue history, prior actions and observations, and available tool instructions and resources \citep{yao2023react}. Through tool calls, the agent can inspect data, write and run scripts, and create report files; the environment returns execution feedback $o_{imt}$ (such as printed outputs, generated files, or error messages), which is added to the context for the next step. DSArena records the full action--feedback sequence in an execution log $\mathcal L_{im}$. Together with the final report $\mathcal R_{im}$, source code $\mathcal C_{im}$, and generated files $\mathcal F_{im}$, this log forms the complete analysis $O_{im}=(\mathcal R_{im},\mathcal C_{im},\mathcal F_{im},\mathcal L_{im})$.

To reduce model-name and position bias, the platform randomly assigns the two competing models to display positions A and B (denoted by identities $a_i$ and $b_i$) and hides their names until voting is complete. Users inspect both outputs without knowing which side corresponds to the model they selected, and their choice is recorded as $y_i=1$ if position A is preferred and $y_i=0$ if position B is preferred.

Because a binary vote shows \emph{which} analysis won without explaining \emph{why}, DSArena also collects optional voting reasons. Users may choose one or more predefined tags or submit a vote without tags. For each reason category $c$, the binary indicator $r_{ic}$ equals $1$ when category $c$ is recorded for the winning analysis in comparison $i$, and $0$ otherwise. The seven available options are ``More accurate'', ``More complete'', ``Better report'', ``Better charts'', ``Better code'', ``Faster'', and ``Other'' (which allows a free-text explanation).

\subsection{Bradley--Terry preference estimation}
\label{sec:bt}
To combine pairwise outcomes across different matchups into a common scale of relative capability, we fit a Bradley--Terry (BT) model \citep{bradley1952rank}. Each agent $m$ has a log-strength $\beta_m\in\R$, corresponding to positive strength $s_m=e^{\beta_m}$. Stacking these parameters into $\beta=(\beta_1,\ldots,\beta_M)^\top$, the probability that agent $a_i$ (in position A) beats agent $b_i$ (in position B) in comparison $i$ is
\begin{equation*}
P_i(\beta)=\Pr(y_i=1\mid a_i,b_i)
=\frac{e^{\beta_{a_i}}}{e^{\beta_{a_i}}+e^{\beta_{b_i}}}
=\sigma(\beta_{a_i}-\beta_{b_i}),
\end{equation*}
where $\sigma(u)=1/(1+e^{-u})$ is the logistic function. A larger positive log-strength gap yields a higher predicted win probability for A, whereas equal log-strengths yield a win probability of $0.5$.

Given any comparison set $\mathcal J$, we estimate $\beta$ by minimizing the ridge-penalized negative log-likelihood subject to a zero-sum centering constraint:
\begin{equation}
\hat\beta_{\mathcal J}=\argmin_{\sum_m\beta_m=0}
\left\{\sum_{i\in\mathcal J}\ell_i(\beta)+\frac{\lambda}{2}\|\beta\|_2^2\right\}.
\label{eq:bt-fit}
\end{equation}
Here $\ell_i(\beta)=-y_i\log P_i(\beta)-(1-y_i)\log[1-P_i(\beta)]$ is the binary cross-entropy loss for comparison $i$, with each comparison contributing one vote. The ridge penalty $\lambda>0$ stabilizes estimates for agents with fewer comparisons or highly one-sided win--loss records, while the zero-sum constraint fixes the arbitrary shift of log-strengths. Because the penalized objective is strictly convex, we find its unique global minimum using Newton's method with backtracking line search \citep{boyd2004convex}.

\subsection{Leaderboard construction}
\label{sec:uncertainty}
Fitting the BT model to all $n$ comparisons yields the full-data log-strength vector $\hat\beta$, which we linearly transform into an interpretable Elo-like rating scale:
\begin{equation}
S_m=\mu_S+\frac{\Delta_S}{\ln 10}(\hat\beta_m-\bar\beta).
\label{eq:rating}
\end{equation}
Here $\bar\beta=M^{-1}\sum_m\hat\beta_m$ is the mean fitted log-strength, $\mu_S$ sets the center of the scale, $\Delta_S>0$ sets the spread (so that a $\Delta_S$-point lead corresponds to 10:1 winning odds; see Appendix~\ref{app:rating-conversion}), and $\ln(\cdot)$ denotes the natural logarithm. Because this mapping is strictly increasing, it preserves the exact BT ordering. Alongside these opponent-adjusted ratings, we report each agent's comparison count $N_m$, win count $W_m$, and raw win rate $\widehat{\mathrm{WR}}_m=W_m/N_m$. To measure sampling uncertainty, we compute 95\% bootstrap confidence intervals (95\% bootstrap CIs) by resampling comparisons with replacement and refitting the BT model across 500 resamples \citep{diciccio1996bootstrap}.

\subsection{Subjective voting reason analysis}
\label{sec:aspects}
While the overall leaderboard ranks models globally, the recorded voting reasons provide a subjective view of \emph{what qualities} drive those wins. First, we count reason frequencies across annotated comparisons, counting each category once per comparison (comparisons with multiple tags contribute to each tagged category). Second, to examine whether different models lead on different dimensions, we select each reason subset $\mathcal J_c=\{i:r_{ic}=1\}$ and refit the BT model under the same objective and ridge penalty $\lambda$. As with the overall leaderboard, we compute 95\% bootstrap CIs for each subset using 300 resamples.

\subsection{Objective execution and report feature analysis}
\label{sec:feature-associations}
Voting reasons reflect subjective user judgments. To measure objective differences between the two analyses and how they relate to preference, we examine nine execution and report features: duration, LLM turns, tool calls, tokens, generated files, report pages, words, figures, and tables. We begin with two descriptive summaries. At the model level, we compute each agent's median value for every feature and measure its rank correlation with overall BT ratings via Spearman's $\rho$. At the comparison level, we compute the raw win rate of the analysis with the larger value for each feature (excluding tied pairs) along with its 95\% bootstrap CI.

However, raw pairwise win rates can be confounded by model identity: a stronger model may win mainly because of better reasoning while also tending to run longer or include more figures. To separate feature associations from baseline differences between competing models, we extend the pairwise BT model into a covariate-adjusted logistic regression. For comparison $i$ and feature $j$, let $q_{i\,a_i\,j}$ and $q_{i\,b_i\,j}$ denote the measured feature values for the agents in positions A and B, respectively. Because execution and report measurements are non-negative and right-skewed, we apply a $\log(1+q)$ transformation and define the standardized A-minus-B feature difference as
\begin{equation*}
z_{ij}=\frac{\log(1+q_{i\,a_i\,j})-\log(1+q_{i\,b_i\,j})}{s_j},
\end{equation*}
where $s_j$ is the standard deviation of the unscaled log-differences across all comparisons. Scaling by $s_j$ without mean-centering places all nine features on a common one-standard-deviation scale while preserving $z_{ij}=0$ whenever the two analyses have equal values.

Using these standardized differences, we fit two complementary logistic regression models. First, to evaluate each feature's marginal association with winning after controlling for the competing models, we fit a \emph{separate} (single-feature) model for each feature $j$:
\begin{equation*}
\Pr(y_i=1\mid a_i,b_i,z_{ij})
=\sigma\!\left(\delta+\eta_{a_i}-\eta_{b_i}+\alpha_j z_{ij}\right).
\end{equation*}
Here $\delta$ is a global intercept capturing any residual position-A preference, {and $\eta_m$ is a model-specific baseline strength parameter estimated jointly in the regression (with one reference agent fixed at $\eta_m=0$ for identifiability), so that $\eta_{a_i}-\eta_{b_i}$ controls for baseline differences between the two competing models.} Second, because execution effort and report content often vary together (for example, longer runs tend to involve more tool calls, words, and figures), we fit a \emph{joint} (multi-feature) model that includes all nine features at once:
\begin{equation*}
\Pr(y_i=1\mid a_i,b_i,z_{i1},\ldots,z_{i9})
=\sigma\!\left(\delta+\eta_{a_i}-\eta_{b_i}+\sum_{j=1}^{9}\alpha_j z_{ij}\right).
\end{equation*}
In this joint model, $\alpha_j$ isolates the conditional association of feature $j$ with user preference while holding both model identities and the other eight features fixed. For both models, we report the odds ratio $\mathrm{OR}_j=e^{\alpha_j}$, which gives the multiplicative change in A's odds of winning per one-standard-deviation increase in $z_{ij}$; values above $1$ indicate a positive association with preference. For both the raw win rates and the regression odds ratios, we compute 95\% bootstrap CIs by resampling comparisons with replacement (and refitting each logistic regression) across 2,000 resamples. Additional measurement and regression details appear in Appendix~\ref{app:feature-models}.

\subsection{Human--LLM judge preference comparison}
\label{sec:judge-comparison}

To examine how closely automated evaluators match real users, we ask three LLM judges to evaluate 500 sampled analysis pairs under the same blinded voting format as the online platform. For each pair, each judge selects its preferred analysis, assigns voting reasons, and writes a detailed explanation. We first compute how often each judge agrees with the corresponding human vote. Then, to see which measurable properties accompany their disagreements, we compare human and LLM-judge selection patterns across the same nine execution and report features.

Specifically, for each feature $j$, we define a directional feature advantage (a higher count or volume for the eight output and effort features, and a \emph{shorter} runtime for duration) and measure how often each evaluator chooses the advantaged analysis. Let $n_j$ be the number of eligible (non-tied) comparisons for feature $j$, and let $c_{ij}^{(s)}\in\{0,1\}$ indicate whether evaluator $s$ chooses the advantaged analysis in comparison $i$. The selection rate and the human--judge gap are defined as
\begin{equation*}
p_j^{(s)}=\frac{1}{n_j}\sum_{i=1}^{n_j}c_{ij}^{(s)},
\qquad
\Delta p_j=100\left[
p_j^{(\mathrm H)}
-\frac{1}{3}\sum_{k=1}^{3}p_j^{(\mathrm L_k)}
\right],
\end{equation*}
where $\mathrm H$ denotes human users and $\mathrm L_k$ denotes LLM judge $k$. Thus, $\Delta p_j$ gives the human selection rate minus the three-judge mean in percentage points: positive values indicate that humans favor the named advantage more strongly than the LLM judges.

Because binary selection rates do not capture the size of feature differences, we also compute the average standardized log-feature margin between the chosen analysis and its alternative. For each eligible comparison, we compute
\begin{equation*}
u_{ij}^{(s)}
=
\log\!\left(1+q_{ij,\mathrm{chosen}}^{(s)}\right)
-
\log\!\left(1+q_{ij,\mathrm{alternative}}^{(s)}\right),
\qquad
M_j^{(s)}
=
\frac{1}{n_j}\sum_{i=1}^{n_j}
\frac{u_{ij}^{(s)}}{\sigma_j},
\end{equation*}
where $q$ denotes the measured feature value and $\sigma_j$ is the standard deviation of the A-minus-B log-feature differences across the $n_j$ eligible comparisons. To match the speed direction above, we reverse the sign of $u_{ij}^{(s)}$ for duration so that positive values denote faster chosen analyses. Accordingly, a positive $M_j^{(s)}$ means that evaluator $s$'s chosen analyses tend to have larger values of feature $j$ (or shorter runtimes for duration) than their alternatives. We estimate 95\% bootstrap CIs via 5,000 paired resamples, resampling comparisons with replacement while keeping human and all three judges' votes together in each resample.

Finally, to see whether pairwise disagreements change the overall model rankings, we fit separate BT leaderboards to the human votes and to each LLM judge's votes over the same 500 comparisons, and compute Spearman's rank correlation between each judge's leaderboard and the human leaderboard.

\section{Data Collection}
\label{sec:data-collection}
Before adding models to DSArena, we screen candidate LLMs to verify that they can reliably run end-to-end data analysis workflows inside LAMBDA. Screening checks multi-step execution reliability, effective use of LAMBDA's long-term memory, instruction following, and stability during internal testing. Only models that consistently complete workflows and produce valid analytical reports are included in the arena. Appendix~\ref{app:model-configurations} lists the configurations of all 12 deployed LLMs.

Over four months, DSArena collected 2,038 pairwise preference votes from 779 users across these 12 LLMs. Each record pairs two models on the same user-submitted task and logs a single blinded human vote, yielding 4,076 model appearances across all 66 unordered model pairs. Alongside their vote, users may optionally select one or more reason tags (``More accurate'', ``More complete'', ``Better report'', ``Better charts'', ``Better code'', ``Faster'', and ``Other''), giving 1,751 comparisons with recorded reason labels.

\begin{figure}[!htbp]
  \centering
  \includegraphics[width=\linewidth]{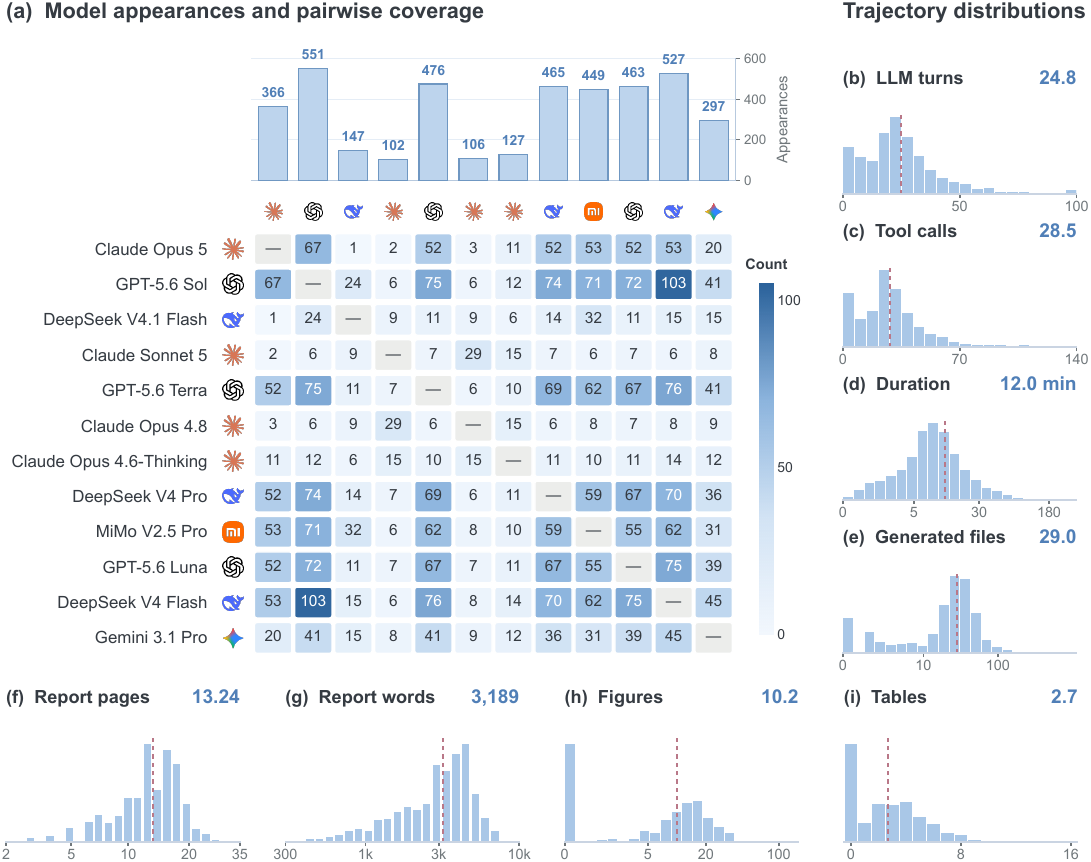}
  \caption{Overview of the DSArena dataset. (a) Comparison counts across all 66 model pairs (heatmap) and total appearances per model (top bars). Panels (b--e) show distributions of LLM turns, tool calls, duration in minutes, and generated files; panels (f--i) show report pages, words, figures, and tables. Dashed lines and labels indicate arithmetic means. Duration, files, pages, words, and figures are plotted on a $\log(1+x)$ scale with tick labels in original units.}
  \label{fig:battles}
\end{figure}

Figure~\ref{fig:battles} summarizes pairwise matchup coverage along with the distributions of execution and report features. The heatmap in panel (a) shows comparison counts across all model pairs, while the bar chart above it shows total appearances per model. Panels (b--i) report the distributions of LLM turns, tool calls, execution duration, generated files, report pages, word counts, figures, and tables. Across execution logs, analyses average 24.8 LLM turns, 28.5 tool calls, 12.0 minutes of runtime, and 10.2 figures generated; reports in PDF format average 13.24 pages and 3,189 words.

\section{Experiments}
\label{sec:setting}
We organize our experiments around four research questions: which LLMs users prefer most (RQ1), what subjective reasons explain those choices (RQ2), which objective execution and report features are associated with winning (RQ3), and how human preferences differ from LLM judgments (RQ4). Below we specify the experimental settings for each analysis following the methods in Section~\ref{sec:methods}.

\subsection{RQ1: Which LLM do users prefer most?}
\label{sec:bt-implementation}
We fit the regularized BT model to all 2,038 human votes with ridge coefficient $\lambda=0.5$ and map log-strengths onto the rating scale using $\mu_S=1500$ and $\Delta_S=400$. Alongside opponent-adjusted ratings, we compute raw win rates and 95\% bootstrap CIs. These settings apply to all subsequent BT fits.

\subsection{RQ2: What subjective reasons are recorded for user preferences?}
To analyze subjective drivers of preference, we count each recorded reason category once per comparison across the 1,751 annotated comparisons. We then fit a separate conditional BT ranking on each reason subset $\mathcal J_c$ following Section~\ref{sec:aspects}, allowing direct comparison of model strengths across specific evaluation dimensions.

\subsection{RQ3: Which objective execution and report features are associated with winning?}
To examine objective factors associated with winning, we first compute Spearman's $\rho$ across the 12 agents between each agent's median value on each of the nine features and its overall BT rating. Next, across all 2,038 comparisons, we compute the observed win rate of the analysis with the larger feature value (excluding tied pairs). Finally, we fit both the separate-feature and joint logistic regression models from Section~\ref{sec:feature-associations} to estimate odds ratios and their 95\% bootstrap CIs adjusted for model identity.

\subsection{RQ4: How do human and LLM-judge preferences differ?}
We sample 500 comparisons and ask three LLM judges (GPT-5.6 Luna, Claude Sonnet 5.5, and MiMo V2.6 Pro) to evaluate each anonymized pair, cast a blinded vote, and explain their choices. First, we measure each judge's pairwise agreement with the corresponding human vote. Next, following Section~\ref{sec:judge-comparison}, we compute feature selection rates, human-minus-judge-mean gaps $\Delta p_j$, and average standardized log-feature margins $M_j^{(s)}$ along with their 95\% bootstrap CIs. To examine direct trade-offs between presentation and accuracy, we also inspect choices on the subset of comparisons where the judges agree that figure quality and correctness favor opposing reports. Finally, we fit separate BT leaderboards to human votes and to each judge's votes over the same 500 comparisons, and compute Spearman's rank correlation between each judge's ranking and the human ranking.

\section{Results}
\subsection{RQ1: Which LLM do users prefer most?}
\label{sec:rq1}
\begin{figure}[!htbp]
  \centering
  \includegraphics[width=\linewidth]{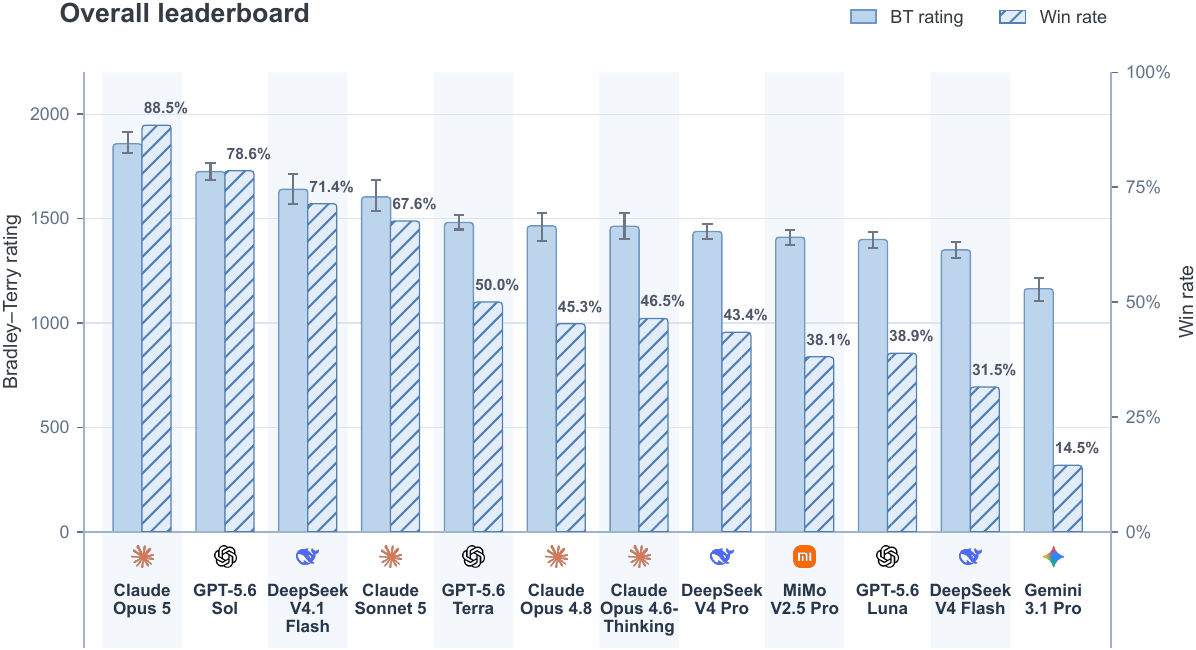}
  \caption{Overall BT ratings and raw win rates across all 2,038 human votes, ordered by BT rating. Solid blue bars represent BT ratings (left axis) and diagonally hatched bars represent raw win rates (right axis); error bars denote 95\% bootstrap CIs.}
  \label{fig:leaderboard}
\end{figure}

Figure~\ref{fig:leaderboard} presents the overall Bradley--Terry leaderboard across all 2,038 human votes, with full numerical results reported in Appendix~\ref{app:overall-leaderboard}. Claude Opus 5 ranks first by a clear margin with a rating of 1,858.2 (winning 324 of 366 comparisons, or 88.5\%), followed by GPT-5.6 Sol at 1,725.1 (433 of 551 comparisons, or 78.6\%) and DeepSeek V4.1 Flash at 1,640.2. Claude Sonnet 5, GPT-5.6 Terra, and Claude Opus 4.8 complete the top six.

The gap between the top two models is 133.1 rating points, with a 95\% bootstrap CI of $[72.5,195.6]$, showing a clear lead for Claude Opus 5 (about 2.15:1 fitted odds; Appendix~\ref{app:rating-conversion}). To understand what drives these preferences, we next examine users' subjective voting reasons (RQ2) and objective execution and report features (RQ3).

\subsection{RQ2: What subjective reasons are recorded for user preferences?}
\label{sec:rq2}

\begin{figure}[!htbp]
\centering
\includegraphics[width=\linewidth]{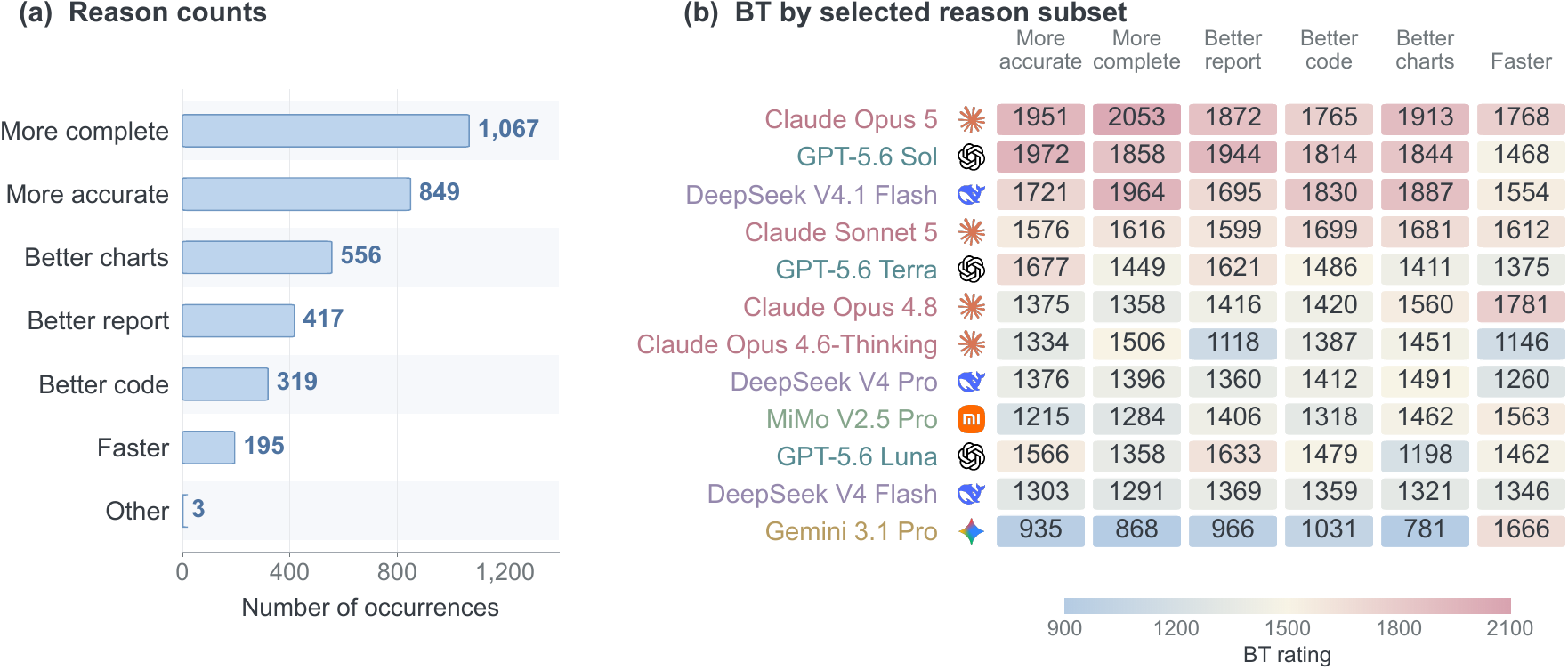}
\caption{Reason frequencies and conditional BT rankings. (a) Number of comparisons carrying each reason tag among annotated comparisons; (b) conditional BT ratings fitted separately to six labeled subsets, ordered by overall BT rating.}
\label{fig:reasons}\label{fig:dimensions}
\end{figure}

Among the 1,751 comparisons with recorded reasons, ``More complete'' is cited most frequently (Figure~\ref{fig:reasons}a), appearing in 1,067 comparisons (60.9\%), followed by ``More accurate'' in 849 (48.5\%) and ``Better charts'' in 556 (31.8\%). By contrast, ``Better report'', ``Better code'', and ``Faster'' appear less often, in 23.8\%, 18.2\%, and 11.1\% of annotated comparisons, respectively. These proportions indicate that users readily recognize and reward thoroughness, such as reports containing more analytical sections, clearer conclusions, and more supporting figures, tables, and pages. In a health-expenditure task, for instance, the winning analysis was credited for going beyond a pooled regression to include country-specific models, nonlinear relationships, and residual comparisons. Similarly, in a housing-price task, the recorded explanation highlighted regression diagnostics, comparisons across predictive models, and five-fold cross-validation. Both cases illustrate how broader exploration and systematic checks contribute to perceived completeness.

Breaking down the leaderboard by reason category shows that no single model leads on every dimension (Figure~\ref{fig:dimensions}b). GPT-5.6 Sol achieves the highest conditional BT ratings on the ``More accurate'' (1,972.1) and ``Better report'' (1,944.1) subsets, whereas Claude Opus 5 leads on ``More complete'' and ``Better charts''. Meanwhile, DeepSeek V4.1 Flash and Claude Opus 4.8 top the ``Better code'' and ``Faster'' subsets, respectively. These different category leaders show that the overall ranking reflects a mix of distinct strengths: analytical coverage, accuracy, presentation quality, and speed do not necessarily favor the same model. Because these subjective reasons are closely tied to measurable differences in runtime and report content, we next examine in RQ3 how objective execution and report features relate to winning.

\subsection{RQ3: Which objective execution and report features are associated with winning?}
\label{sec:rq3}

\begin{figure}[!htbp]
  \centering
  \includegraphics[width=\linewidth]{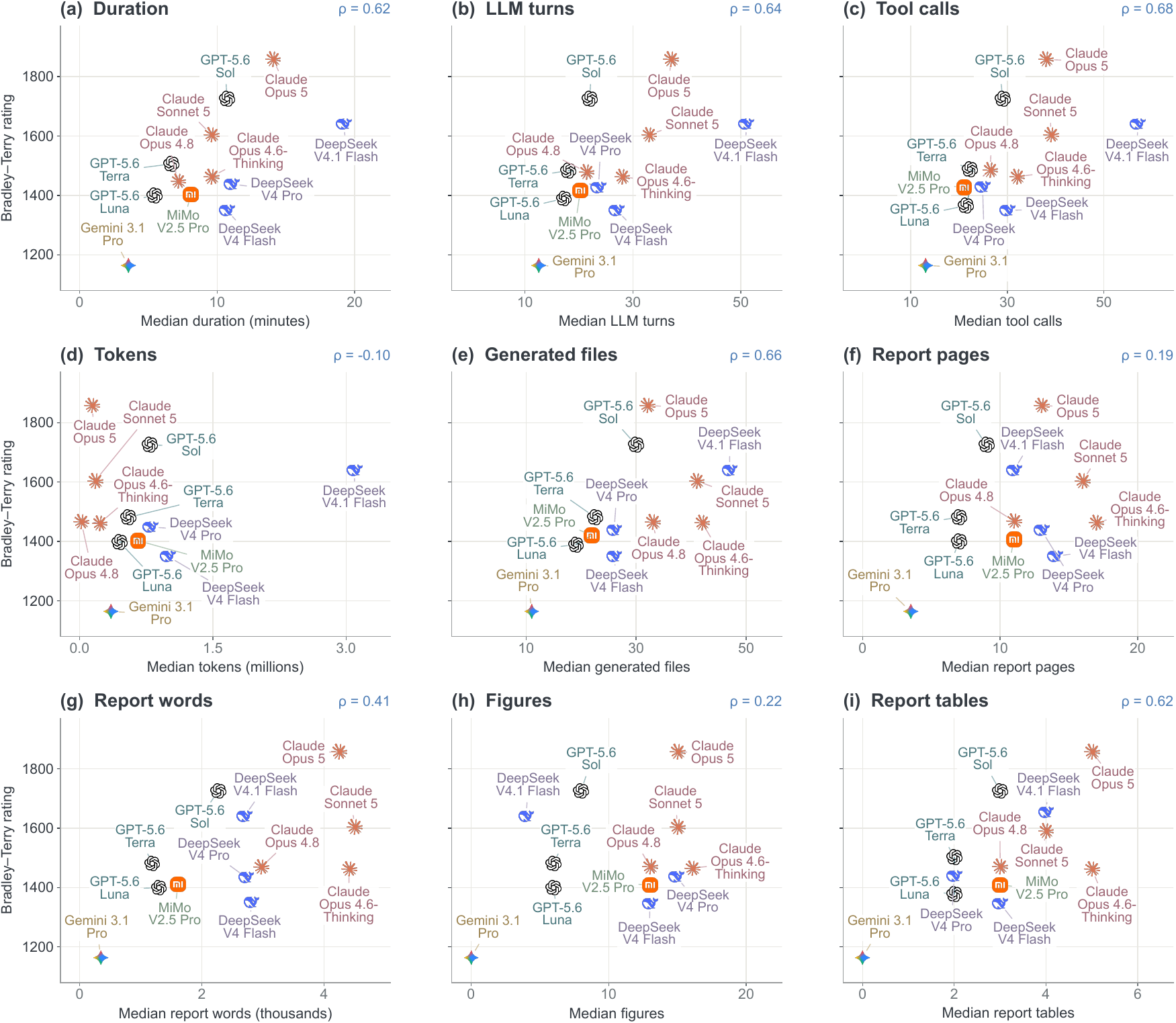}
  \caption{Overall BT rating versus per-agent median values of nine execution and report features across the 12 agents. Positive values of Spearman's $\rho$ indicate that larger median feature values tend to accompany higher BT ratings.}
  \label{fig:latency}
\end{figure}

At the model level, Figure~\ref{fig:latency} plots each agent's overall BT rating against its median values across the nine execution and report features. Among all features, tool calls show the strongest positive rank correlation with model rating ($\rho=0.676$), suggesting that higher-ranked agents generally perform more analytical operations per task. Execution time, however, varies widely even among top-tier models: GPT-5.6 Sol finishes in a median of 10.7 minutes, compared with 14.1 minutes for Claude Opus 5 and 19.2 minutes for DeepSeek V4.1 Flash. Thus, while Claude Opus 5 leads overall human preference, GPT-5.6 Sol offers noticeably faster completion. To move beyond model-level medians, we next analyze how feature differences within individual matchups relate to user choices.

\begin{table}[!htbp]
\centering\small
\caption{Associations between human preferences and execution/report features across all 2,038 comparisons. OR denotes the winning odds ratio per one-standard-deviation increase in the paired log-feature difference. Bold joint-model entries have 95\% bootstrap CIs excluding 1.}
\label{tab:feature-preferences}
\begingroup
\newcommand{\FullCI}[3]{\ensuremath{#1\;[#2,\,#3]}}
\newcommand{\FullRate}[3]{\ensuremath{#1\;[#2,\,#3]}}
\newcommand{\FullSigCI}[3]{\ensuremath{\mathbf{#1}\;[#2,\,#3]}}
\setlength{\tabcolsep}{4pt}
\renewcommand{\arraystretch}{1.33}
\begin{tabular*}{\linewidth}{@{\extracolsep{\fill}}lccc@{}}
\toprule
& \multicolumn{1}{c}{\textbf{Observed choices}}
& \multicolumn{1}{c}{\textbf{Separate model}}
& \multicolumn{1}{c}{\textbf{Joint model}} \\
\cmidrule(lr){2-2}\cmidrule(lr){3-3}\cmidrule(l){4-4}
\textbf{Feature} & \textbf{Win rate (\%) [95\% Bootstrap CI]}
& \textbf{OR [95\% Bootstrap CI]} & \textbf{OR [95\% Bootstrap CI]} \\
\midrule
Longer duration & \FullRate{69.3}{67.2}{71.4} & \FullCI{1.83}{1.56}{2.17} & \FullSigCI{1.59}{1.29}{1.99} \\
More LLM turns & \FullRate{62.4}{60.3}{64.6} & \FullCI{1.51}{1.31}{1.75} & \FullCI{0.99}{0.67}{1.48} \\
More tool calls & \FullRate{64.1}{62.0}{66.3} & \FullCI{1.48}{1.29}{1.72} & \FullCI{1.08}{0.74}{1.55} \\
More tokens & \FullRate{49.7}{47.4}{51.8} & \FullCI{1.56}{1.32}{1.87} & \FullCI{0.94}{0.66}{1.32} \\
More files & \FullRate{67.2}{64.9}{69.3} & \FullCI{1.48}{1.25}{1.82} & \FullSigCI{1.24}{1.01}{1.57} \\
More pages & \FullRate{50.4}{48.0}{52.8} & \FullCI{1.31}{1.14}{1.55} & \FullCI{0.73}{0.45}{1.14} \\
More words & \FullRate{68.3}{66.2}{70.4} & \FullCI{1.28}{1.11}{1.52} & \FullCI{0.81}{0.52}{1.26} \\
More figures & \FullRate{57.2}{54.9}{59.5} & \FullCI{1.51}{1.29}{1.83} & \FullSigCI{1.68}{1.36}{2.14} \\
More tables & \FullRate{68.7}{66.4}{70.9} & \FullCI{1.68}{1.46}{1.96} & \FullSigCI{1.58}{1.34}{1.91} \\
\bottomrule
\end{tabular*}
\endgroup

\end{table}

At the comparison level, Table~\ref{tab:feature-preferences} summarizes raw win rates alongside the separate-feature and joint logistic regression estimates across all 2,038 comparisons. In raw counts, analyses with longer durations win 69.3\% of eligible matchups, while those containing more tables or more words win 68.7\% and 68.3\%, respectively. When each feature is entered individually into a model-adjusted regression, all nine features show positive associations with winning, with 95\% bootstrap CIs strictly above 1.

Because more thorough exploration increases execution effort and report length together, these single-feature associations are strongly correlated. Entering all nine features into the joint model separates their independent effects (Table~\ref{tab:feature-preferences}): execution duration ($\mathrm{OR}=1.59$, 95\% bootstrap CI $[1.29,1.99]$), figure count ($\mathrm{OR}=1.68$, 95\% bootstrap CI $[1.36,2.14]$), and table count ($\mathrm{OR}=1.58$, 95\% bootstrap CI $[1.34,1.91]$) show the largest independent positive associations, while generated files show a smaller positive association whose 95\% bootstrap CI narrowly excludes 1 ($\mathrm{OR}=1.24$, $[1.01,1.57]$). In contrast, the 95\% bootstrap CIs for LLM turns, tool calls, tokens, pages, and words all include 1. Thus, the apparent preference for more turns or longer prose is largely absorbed by correlated output features; what independently predicts user preference is sustained analytical effort paired with rich visual, tabular, and file outputs.

Taken together, the objective results in RQ3 align closely with the subjective voting reasons in RQ2. First, the positive association for execution duration matches users citing ``More complete'' most frequently (60.9\%) and ``Faster'' least frequently (11.1\%): across completed comparisons, users generally value deeper analysis over shorter runtime. Second, the strong independent effects of figures and tables correspond directly to the frequent selection of ``Better charts'' (31.8\%) and ``Better report'' (23.8\%), whereas the lack of independent effects for LLM turns, tool calls, tokens, and word count is consistent with users placing less emphasis on internal code steps (``Better code'', 18.2\%) or raw text length alone. This alignment also explains the top of the leaderboard: Claude Opus 5 leads overall because it excels in ``More complete'' and ``Better charts'', the exact qualities backed by both subjective feedback and objective feature associations. At the same time, objective counts cannot directly measure analytical correctness (``More accurate'', 48.5\%); RQ4 therefore compares human choices against LLM judges to examine how presentation and statistical rigor are weighed in practice.

\subsection{RQ4: How do human and LLM-judge preferences differ?}
\label{sec:rq4}
\begin{figure}[!htbp]
  \centering
  \includegraphics[width=\linewidth]{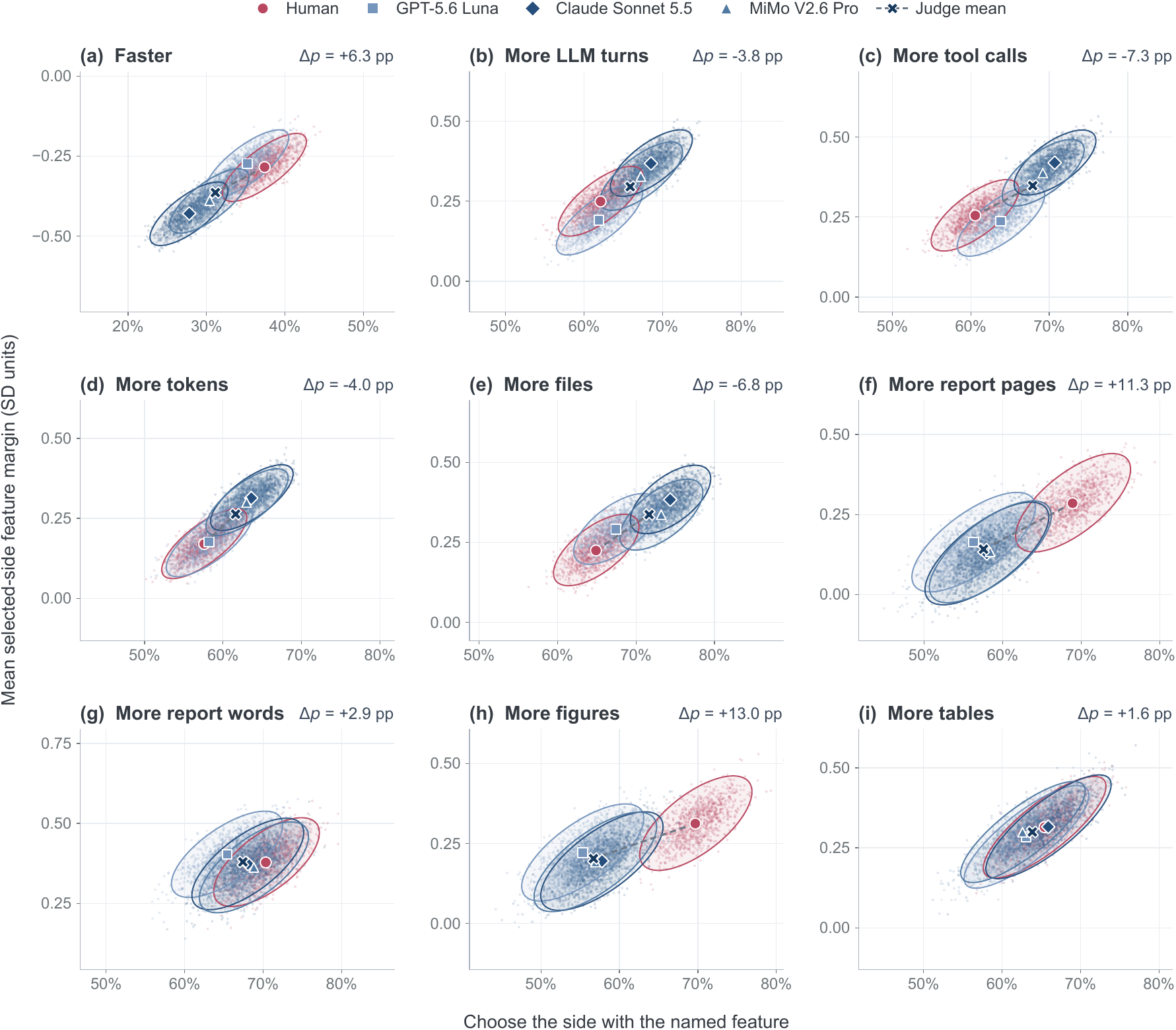}
  \caption{Human versus LLM-judge preference profiles across nine features on 500 sampled comparisons. Horizontal axes show selection rates for the named feature advantage and vertical axes show mean standardized log-feature margins ($M_j^{(s)}$); ellipses indicate 95\% bootstrap regions. Dashed lines connect the human profile to the three-judge mean, annotated with the selection-rate gap $\Delta p$ in percentage points.}
  \label{fig:judge-preferences}
\end{figure}

Across the 500 sampled comparisons, all three LLM judges show moderate and consistent pairwise agreement with human votes: 66.2\% for GPT-5.6 Luna, 67.2\% for Claude Sonnet 5.5, and 66.6\% for MiMo V2.6 Pro. To explain the remaining roughly one-third of disagreements, Figure~\ref{fig:judge-preferences} compares human and LLM-judge selection profiles across the nine execution and report features.

The largest human-minus-judge selection-rate gaps appear for figure count and report pages (panels h and f). Specifically, humans choose the report with more figures in 69.7\% of eligible comparisons versus a three-judge mean of 56.7\% ($\Delta p=+13.0$ percentage points; 95\% bootstrap CI $[6.5,19.6]$). Likewise, humans select the report with more pages in 68.9\% of comparisons versus a judge mean of 57.6\% ($\Delta p=+11.3$ points; 95\% bootstrap CI $[4.9,18.3]$). Consistent with these selection rates, human-chosen reports also have larger standardized figure-count and page-count margins over their alternatives (vertical axis $M_j^{(s)}$ in Figure~\ref{fig:judge-preferences}). By contrast, the human--judge gaps for word count ($+2.9$ points) and table count ($+1.6$ points) are small. Thus, humans do not simply prefer longer reports across every measure of size; rather, visual content and report length in pages form a clear source of disagreement whose 95\% bootstrap CIs lie well above zero.

A contrasting pattern appears for execution speed and tool usage. Although both humans and LLM judges choose the faster analysis less than half the time (reflecting a shared preference for deeper work), humans choose the faster run more often than judges do: 37.4\% versus a three-judge mean of 31.1\% ($\Delta p=+6.3$ points; Figure~\ref{fig:judge-preferences}a). Conversely, for tool calls and generated files (panels c and e), LLM judges show stronger preferences than humans: human selection rates are 60.6\% and 64.9\%, compared with judge means of 67.9\% and 71.7\% ($\Delta p=-7.3$ and $-6.8$ points, respectively), along with larger judge-selected feature margins. Overall, compared with the judge average, humans lean more toward faster completion and visually richer reports, whereas LLM judges more often favor analyses with more tool calls and generated files.

This difference becomes even clearer when presentation and methodological correctness conflict directly. Across the 37 comparisons where the judges agree that figure quality and correctness favor opposite reports, GPT-5.6 Luna, Claude Sonnet 5.5, and MiMo V2.6 Pro pick the higher-correctness report in 78.4\%, 67.6\%, and 78.4\% of cases, respectively, whereas humans do so in only 40.5\% of cases. Reading the judges' written explanations links this gap directly to statistical rigor and the reliability of conclusions. In one survival-analysis comparison, for example, the human user preferred Claude Opus 4.6-Thinking, while all three judges selected GPT-5.6 Terra; Claude Sonnet 5.5 and MiMo V2.6 Pro explicitly favored censoring-aware inference and properly distinguishing a non-significant result from evidence of equivalence, even though the competing report had stronger charts. Together, these results show that LLM judges weigh statistical justification more heavily, while human users place relatively greater weight on visual clarity and presentation.

\begin{figure}[!htbp]
  \centering
  \includegraphics[width=\linewidth]{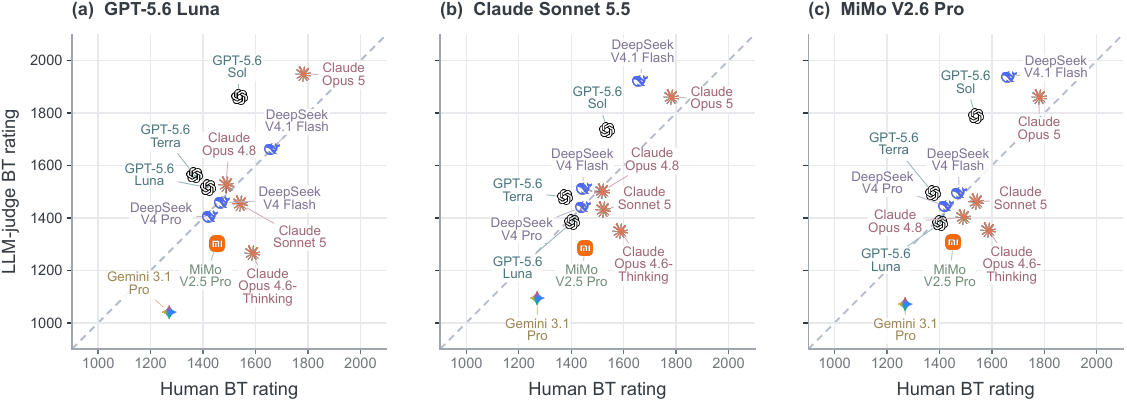}
  \caption{Comparison of human and LLM-judge BT rankings on the sampled 500 comparisons. Each panel plots human BT ratings (horizontal axis) against one judge's BT ratings (vertical axis), with model names linked to their family logos. The dashed diagonal indicates equal ratings.}
  \label{fig:judge-ranks}
\end{figure}

Importantly, these pairwise disagreements also change the resulting model rankings. Figure~\ref{fig:judge-ranks} compares the BT leaderboard estimated from human votes with those estimated from each LLM judge on the same 500 comparisons. Claude Sonnet 5.5 aligns most closely with the human ordering (Spearman $\rho=0.643$), followed by MiMo V2.6 Pro ($\rho=0.538$) and GPT-5.6 Luna ($\rho=0.455$). These weak-to-moderate rank correlations indicate that human and LLM-judge rankings align only loosely, so replacing human voters with LLM judges alters not only individual comparison outcomes, but also the overall ranking of data science agents.

\section{Discussion}
\subsection{Preference differences between humans and LLM judges}
The systematic gaps between human and LLM-judge preferences may reflect differences in how the two groups read and assess analytical reports. Users evaluating analyses of their own problems often look for key findings quickly rather than conducting a full methodological check. In this setting, informative figures and structured tables make distributions, comparisons, and main results much easier to inspect than dense text or isolated numbers, making clear visual presentation central to communicating analytical value.

LLM judges, by contrast, emphasize validation checks, numerical consistency, uncertainty estimates, and tight alignment between statistical evidence and conclusions. Their written explanations reflect a systematic focus on formal justification and reproducibility, favoring reports that explicitly state confidence intervals, diagnostic checks, limitations, and the conditions under which conclusions hold. While these checks are essential for analytical reliability, an automated rubric that heavily rewards explicit limitations and qualifications can undervalue exploratory insight, effective modeling choices, and clear visual communication. Human preferences therefore provide an important complement to automated evaluation, capturing whether an analysis clearly communicates its findings and delivers practical value to the user.

\subsection{Balancing analytical depth and timely delivery}
While Sections~\ref{sec:rq2} and~\ref{sec:rq3} show that users reward deeper exploration and richer reports, their patience is not unlimited: platform logs indicate that when users manually stopped an unfinished run after the competing analysis had already finished, 80.6\% of stops occurred within ten minutes. Taken together, these observations highlight a practical trade-off between analytical depth and timely delivery.

This trade-off motivates designing data science agents for staged delivery. Rather than holding back all results until a long run finishes, an agent could first deliver a coherent initial report covering the main findings, and then continue exploring other approaches, running additional checks, and updating the report over time. Users could inspect the initial findings right away and decide whether to wait for deeper analysis, redirect the exploration, or stop early. Such a design offers earlier access to useful results while keeping the option for more thorough analysis.

\subsection{Reward modeling}
Beyond benchmarking, DSArena's pairwise comparisons provide real-world preference data for training reward models over complete data science workflows. Given a user task and two candidate analyses, a reward model can learn to score full outputs (including reports, source code, and generated files) so that user-preferred analyses receive higher rewards \citep{christiano2017preferences,stiennon2020summarize,ouyang2022instructions}. Because our RQ4 results show that off-the-shelf LLM judges differ from human users in systematic ways, reward models trained directly on human preferences offer a promising path for aligning data science agents with real user needs. Alternatively, reward modeling can combine human preferences (capturing practical utility and presentation clarity) with automated checks or LLM-judge signals (capturing statistical rigor and methodological correctness), balancing user-facing quality with analytical reliability.

\section{Conclusion}
We presented DSArena, an online platform built on LAMBDA that evaluates LLM-based data science agents through real human preferences. By letting users submit their own tasks and datasets and judge complete analyses via blinded pairwise voting, DSArena captures practical analytical value while reducing benchmark contamination. Across 2,038 votes from 779 users covering 12 LLMs over four months, Claude Opus 5 leads the Bradley--Terry leaderboard, followed by GPT-5.6 Sol and DeepSeek V4.1 Flash. Beyond ranking models, our subjective and objective preference analyses show that users strongly value analytical completeness alongside clear visual and tabular presentation: ``More complete'' is the most frequently cited voting reason, and execution duration, figure count, and table count each remain independently associated with winning even after controlling for model identity and correlated features. Comparing human choices with three LLM judges across 500 sampled comparisons also reveals systematic differences: humans favor visually richer reports more often, whereas LLM judges reward heavier tool use, formal validation, and statistical support, yielding only moderate rank agreement with human leaderboards. In practice, balancing thoroughness against waiting time also points toward staged-delivery agent designs. Overall, DSArena complements automated benchmarks with the perspective of real users and provides human preference data for future reward modeling and agent alignment.

\bibliographystyle{unsrtnat}
\IfFileExists{DSArena/dsarena_references.bib}{%
  \bibliography{DSArena/dsarena_references}%
}{%
  \bibliography{dsarena_references}%
}

\clearpage
\appendix
\section*{Appendix}
\section{Notation}
\label{app:notation}
Table~\ref{tab:notation} summarizes the notation used throughout the methods and supplementary derivations.
\begin{table}[H]
\centering\small\renewcommand{\arraystretch}{1.15}
\caption{Notation for tasks, agents, human preferences, and statistical analysis.}
\label{tab:notation}
\begin{tabularx}{\linewidth}{@{}p{0.40\linewidth}X@{}}\toprule
\textbf{Symbol} & \textbf{Meaning} \\ \midrule
$n,M$; $i,m,t,j,c$ & Numbers of comparisons and agents; indices for comparison, agent, execution step, feature, and reason. \\
$z_i=(p_i,\mathcal D_i)$ & Task $i$: user instruction $p_i$ and uploaded data or files $\mathcal D_i$. \\
$\theta_m,\pi_{\theta_m}$ & LLM parameters and conditional policy over actions. \\
$H_{imt}$; $v_{imt},o_{imt}$; $\mathcal L_{im}$ & Context; action and returned feedback at step $t$; ordered execution log. \\
$O_{im}=(\mathcal R_{im},\mathcal C_{im},\mathcal F_{im},\mathcal L_{im})$ & Complete analysis: report, code, generated files, and execution log. \\
$a_i,b_i,y_i$ & Model identities in positions A and B, and binary choice indicator ($y_i=1$ if A wins). \\
$r_{ic}$; $\mathcal J,\mathcal J_c$ & Binary indicator for reason $c$; comparison set and subset carrying reason $c$. \\
$\beta_m,\beta,s_m$; $\hat\beta_{\mathcal J},\hat\beta,\bar\beta$ & Log-strength, log-strength vector, and positive strength $s_m=e^{\beta_m}$; subset fit, full-data fit, and fitted mean. \\
$P_i(\beta),\sigma$; $\ell_i,\lambda$ & BT A-win probability and logistic function; per-comparison loss and ridge penalty coefficient. \\
$S_m$; $\mu_S,\Delta_S$ & Leaderboard rating; mean rating center and positive scale parameter. \\
$N_m,W_m,\widehat{\mathrm{WR}}_m$ & Model comparison count, win count, and raw win rate. \\
$q_{imj}$; $z_{ij},s_j$ & Feature measurement; standardized A-minus-B log-feature difference and its unscaled standard deviation. \\
$\delta,\eta_m,\alpha_j$; $\rho$ & Position-A intercept, agent coefficient, and feature coefficient in the logistic regression; Spearman correlation. \\
$n_j, c_{ij}^{(s)}, p_j^{(s)}, \Delta p_j$ & Eligible comparison count, feature-advantage indicator, selection rate for evaluator $s$, and human--judge gap (pp). \\
$u_{ij}^{(s)}, M_j^{(s)}, \sigma_j$ & Chosen-minus-alternative log-feature difference, mean standardized margin, and subset standard deviation. \\
\bottomrule
\end{tabularx}
\end{table}

\section{Statistical methods}

\subsection{Deriving win probabilities from rating scores}
\label{app:rating-conversion}
Equation~\eqref{eq:rating} maps the fitted BT log-strengths onto the leaderboard rating scale. To recover pairwise win probabilities directly from these ratings, we first take the difference between the two competing agents in comparison $i$:
\begin{align*}
S_{b_i}-S_{a_i}
&=\left[\mu_S+\frac{\Delta_S}{\ln 10}(\hat\beta_{b_i}-\bar\beta)\right]\\
&\quad-\left[\mu_S+\frac{\Delta_S}{\ln 10}(\hat\beta_{a_i}-\bar\beta)\right]\\
&=\frac{\Delta_S}{\ln 10}(\hat\beta_{b_i}-\hat\beta_{a_i}).
\end{align*}
Both the common center $\mu_S$ and the mean log-strength $\bar\beta$ cancel out. Rearranging yields
\begin{equation*}
\hat\beta_{b_i}-\hat\beta_{a_i}
=\frac{\ln 10}{\Delta_S}(S_{b_i}-S_{a_i}).
\end{equation*}
Substituting this relation into the fitted BT probability and using the identity $e^{(\ln 10)u}=10^u$, we obtain
\begin{align*}
P_i(\hat\beta)
&=\frac{1}{1+e^{\hat\beta_{b_i}-\hat\beta_{a_i}}}\\
&=\frac{1}{1+e^{\frac{\ln 10}{\Delta_S}(S_{b_i}-S_{a_i})}}\\
&=\frac{1}{1+10^{(S_{b_i}-S_{a_i})/\Delta_S}}.
\end{align*}
Hence, leaderboard ratings encode the exact same BT probabilities on a base-10 scale: equal ratings correspond to $P_i(\hat\beta)=1/2$, while a $\Delta_S$-point lead for agent A implies
\begin{equation*}
P_i(\hat\beta)=\frac{1}{1+10^{-1}}=\frac{10}{11}\approx0.9091
\end{equation*}
(10:1 winning odds). Under our scale setting $\Delta_S=400$, the 133.1-point lead of Claude Opus 5 over GPT-5.6 Sol corresponds to approximately 2.15:1 fitted odds.

\subsection{Feature measurements and regression models}
\label{app:feature-models}
For comparison $i$ and agent $m$, $q_{imj}$ denotes the measured value of feature $j$. When counting figures, individual diagnostic subplots are counted separately (so a $2\times2$ diagnostic grid contributes four figures), whereas a complete scatter matrix or pairplot (including triangular layouts) counts as a single figure; legends, colorbars, and decorative images are excluded.

In the pairwise preference regressions, each feature enters as the standardized A-minus-B log-difference
\begin{equation*}
z_{ij}=\frac{\log(1+q_{i\,a_i\,j})-\log(1+q_{i\,b_i\,j})}{s_j},
\end{equation*}
where $s_j$ is the population standard deviation of the unscaled log-differences across all 2,038 comparisons. Applying $\log(1+q)$ handles recorded zeros and reduces skewness, while scaling by $s_j$ without subtracting the mean preserves $z_{ij}=0$ whenever both analyses have identical measurements.

Each specification includes a position-A intercept $\delta$ and the model-effect contrast $\eta_{a_i}-\eta_{b_i}$, with one reference model coefficient fixed at zero for identifiability. Separate-feature regressions include a single covariate $\alpha_j z_{ij}$, whereas the joint model includes all nine features simultaneously:
\begin{equation*}
\operatorname{logit}\Pr(y_i=1\mid a_i,b_i,z_{i1},\ldots,z_{i9})
=\delta+(\eta_{a_i}-\eta_{b_i})+\sum_{j=1}^9\alpha_j z_{ij},
\end{equation*}
where $\operatorname{logit}(p)=\log[p/(1-p)]$. We fit unpenalized binomial logistic regressions and report odds ratios as $e^{\alpha_j}$, obtaining 95\% bootstrap CIs from the 2.5th and 97.5th percentiles.

\section{Supplementary tables}
\subsection{Overall leaderboard}
\label{app:overall-leaderboard}
Table~\ref{tab:leaderboard} gives the numerical results underlying Figure~\ref{fig:leaderboard}, including comparison counts, wins, and 95\% bootstrap CIs.
\begin{table}[H]
\centering\small\renewcommand{\arraystretch}{1.15}
\caption{Overall Bradley--Terry leaderboard. Each comparison counts once. Raw win rates describe observed wins; BT ratings account for the opponents compared.}
\label{tab:leaderboard}
\begin{tabular*}{\linewidth}{@{\extracolsep{\fill}}rlrrrrr@{}}\toprule
\textbf{Rank} & \textbf{Model-backed agent} & \textbf{Rating} & \textbf{95\% Bootstrap CI} & \textbf{Battles} & \textbf{Wins} & \textbf{Win rate} \\
\midrule
1 & Claude Opus 5 & 1858.2 & [1812.9, 1912.9] & 366 & 324 & 88.5\% \\
2 & GPT-5.6 Sol & 1725.1 & [1683.3, 1767.6] & 551 & 433 & 78.6\% \\
3 & DeepSeek V4.1 Flash & 1640.2 & [1571.1, 1713.1] & 147 & 105 & 71.4\% \\
4 & Claude Sonnet 5 & 1604.1 & [1537.3, 1683.3] & 102 & 69 & 67.6\% \\
5 & GPT-5.6 Terra & 1481.1 & [1447.8, 1519.0] & 476 & 238 & 50.0\% \\
6 & Claude Opus 4.8 & 1465.0 & [1394.8, 1527.7] & 106 & 48 & 45.3\% \\
7 & Claude Opus 4.6-Thinking & 1463.1 & [1401.4, 1524.8] & 127 & 59 & 46.5\% \\
8 & DeepSeek V4 Pro & 1438.3 & [1404.0, 1473.2] & 465 & 202 & 43.4\% \\
9 & MiMo V2.5 Pro & 1410.8 & [1372.7, 1444.1] & 449 & 171 & 38.1\% \\
10 & GPT-5.6 Luna & 1399.7 & [1360.9, 1433.6] & 463 & 180 & 38.9\% \\
11 & DeepSeek V4 Flash & 1350.3 & [1312.2, 1385.8] & 527 & 166 & 31.5\% \\
12 & Gemini 3.1 Pro & 1164.2 & [1106.9, 1213.4] & 297 & 43 & 14.5\% \\
\bottomrule
\end{tabular*}
\end{table}

\subsection{Model configurations}
\label{app:model-configurations}
\begin{table}[H]
\centering\small\renewcommand{\arraystretch}{1.15}
\caption{Model configurations in DSArena. A dash indicates that no reasoning level is explicitly set.}
\label{tab:model-configurations}
\begin{tabular*}{\linewidth}{@{\extracolsep{\fill}}lcrl@{}}\toprule
\textbf{Model} & \textbf{Reasoning level} & \textbf{Max tokens} & \textbf{Vision support} \\ \midrule
Claude Opus 5 & --- & 32768 & Native \\
GPT-5.6 Sol & High & 32768 & Native \\
DeepSeek V4.1 Flash & --- & 32768 & Native \\
Claude Sonnet 5 & --- & 32768 & Native \\
GPT-5.6 Terra & High & 32768 & Native \\
Claude Opus 4.8 & --- & 32768 & Native \\
Claude Opus 4.6-Thinking & High & 32768 & Native \\
DeepSeek V4 Pro & --- & 32768 & GPT-5.6 Luna (tool) \\
MiMo V2.5 Pro & --- & 32768 & GPT-5.6 Luna (tool) \\
GPT-5.6 Luna & High & 32768 & Native \\
DeepSeek V4 Flash & --- & 32768 & GPT-5.6 Luna (tool) \\
Gemini 3.1 Pro & Low & 32768 & Native \\
\bottomrule
\end{tabular*}
\end{table}

\subsection{Model names and identifiers}
\begin{table}[H]
\centering\small\renewcommand{\arraystretch}{1.15}
\caption{Model names and their recorded identifiers. Figures combine these names with model-family logos.}
\begin{tabular*}{\linewidth}{@{\extracolsep{\fill}}ll@{}}\toprule
\textbf{Display name} & \textbf{Recorded identifier} \\ \midrule
Claude Opus 5 & \texttt{claude-opus-5} \\
GPT-5.6 Sol & \texttt{gpt-5.6-sol} \\
DeepSeek V4.1 Flash & \texttt{deepseek-v4.1-flash} \\
Claude Sonnet 5 & \texttt{claude-sonnet-5} \\
GPT-5.6 Terra & \texttt{gpt-5.6-terra} \\
GPT-5.6 Luna & \texttt{gpt-5.6-luna} \\
Claude Opus 4.8 & \texttt{claude-opus-4-8} \\
Claude Opus 4.6-Thinking & \texttt{claude-opus-4-6-thinking} \\
DeepSeek V4 Pro & \texttt{deepseek-v4-pro} \\
MiMo V2.5 Pro & \texttt{mimo-v2.5-pro} \\
DeepSeek V4 Flash & \texttt{deepseek-v4-flash} \\
Gemini 3.1 Pro & \texttt{gemini-3.1-pro-low} \\
\bottomrule\end{tabular*}
\end{table}
\end{document}